\documentclass[conference]{IEEEtran}
\IEEEoverridecommandlockouts
\usepackage{cite}
\usepackage{amsmath,amssymb,amsfonts}
\usepackage{algorithmic}
\usepackage{algorithm}
\usepackage{graphicx}
\usepackage{textcomp}
\usepackage{xcolor}
\def\BibTeX{{\rm B\kern-.05em{\sc i\kern-.025em b}\kern-.08em
    T\kern-.1667em\lower.7ex\hbox{E}\kern-.125emX}}
\begin{document}

\title{Physics-Enhanced TinyML for Real-Time Detection of Ground Magnetic Anomalies\\

\thanks{This work was supported by NSF EPSCoR Award OIA-1920965.}
}

\author{\IEEEauthorblockN{Talha Siddique}
\IEEEauthorblockA{\textit{Department of Electrical and Computer Engineering} \\
\textit{University of New Hampshire}\\
Durham, NH, USA\\
talha.siddique@unh.edu}
\and
\IEEEauthorblockN{MD Shaad Mahmud}
\IEEEauthorblockA{\textit{Department of Electrical and Computer Engineering} \\
\textit{University of New Hampshire}\\
Durham, NH, USA \\
mdshaad.mahmud@unh.edu}
}

\maketitle

\begin{abstract}
Space weather phenomena like geomagnetic disturbances (GMDs) and geomagnetically induced currents (GICs) pose significant risks to critical technological infrastructure. While traditional predictive models, grounded in simulation, hold theoretical robustness, they grapple with challenges, notably the assimilation of imprecise data and extensive computational complexities. In recent years, Tiny Machine Learning (TinyML) has been adopted to develop Machine Learning (ML)-enabled magnetometer systems for predicting real-time terrestrial magnetic perturbations as a proxy measure for GIC. While TinyML offers efficient, real-time data processing, its intrinsic limitations prevent the utilization of robust methods with high computational needs. This paper developed a physics-guided TinyML framework to address the above challenges. This framework integrates physics-based regularization at the stages of model training and compression, thereby augmenting the reliability of predictions. The developed pruning scheme within the framework harnesses the inherent physical characteristics of the domain, striking a balance between model size and robustness. The study presents empirical results, drawing a comprehensive comparison between the accuracy and reliability of the developed framework and its traditional counterpart. Such a comparative analysis underscores the prospective applicability of the developed framework in conceptualizing robust, ML-enabled magnetometer systems for real-time space weather forecasting.
\end{abstract}


\section{Introduction}

The magnetosphere encompasses the area in space where the Earth’s magnetic field is predominant. Disturbances in the Earth's magnetic field, known as geomagnetic disturbances (GMDs), arise from interactions between the Earth's magnetosphere and the solar wind, a stream of charged particles emanating from the Sun \cite{camporeale_machine_2018} \cite{geosciences12010027}. These geomagnetic disturbances subsequently lead to the generation of geomagnetically induced currents (GICs) within terrestrial conductors \cite{geosciences12010027}\cite{oliveira_geomagnetically_2017}\cite{pirjola_geomagnetically_2000}. GICs pose a significant risk to critical technological infrastructure, including satellite networks and electrical power grids. The severity of this threat was highlighted during the 1989 blackout in Quebec, with projections indicating a potential trillion-dollar global impact from analogous future events \cite{boteler2018dealing}\cite{love2022mapping}\cite{Wang2020}\cite{jonas}\cite{krausmann2016space}. Strategies for mitigating these risks encompass monitoring via space-borne satellites and terrestrial magnetometers and applying predictive models derived from the acquired data \cite{johnson2016review}\cite{freeman2021safety}\cite{krausmann2016space}. Nonetheless, the accurate quantification of GICs is challenging, attributed to proprietary restrictions imposed by energy corporations, necessitating the employment of alternative indicators such as the rate of change in the ground horizontal magnetic component ($dB_{H}/dt$)  \cite{Wintoft2015}\cite{Keesee}\cite{pinto_revisiting_2022}\cite{Siddique2022}. While traditional forecasting methodologies leverage physics-based simulation models, offering theoretical robustness and interpretability, they are constrained by difficulties in assimilating imprecise observational data for determining initial and boundary conditions and their substantial computational complexity \cite{morley2020challenges}\cite{feng2020current}.

Recent efforts have been made to devise a magnetometer system augmented with machine learning (ML) capabilities for instantaneous baseline correction and forecasting of geomagnetic perturbations \cite{Siddique_Sensors}. This endeavor exploits a specialized segment of edge ML known as Tiny Machine Learning (TinyML). TinyML is principally concerned with developing ML models for edge devices with limited resources, such as microcontroller units (MCU) \cite{dutta2021tinyml}\cite{abadade2023comprehensive}\cite{ray2022review}. Initial outcomes from the ML-augmented magnetometer have demonstrated its potential for efficient, low-power, and real-time data processing \cite{Siddique_Sensors}. Nevertheless, due to the intrinsic resource constraints of the intended hardware, the TinyML framework cannot support models with high computational demands \cite{banbury2020benchmarking}\cite{ray2022review}. The TinyML pipeline necessitates converting trained ML models into more lightweight, compressed versions \cite{dutta2021tinyml}\cite{banbury2020benchmarking}. Various model compression schemes and methodologies have been proposed in the literature to address this requirement \cite{dutta2021tinyml}. These techniques generally encompass quantization, pruning, knowledge distillation, and low-rank factorization. Quantization reduces the numerical precision of the model’s parameters, thereby decreasing memory requirements \cite{cheng2017survey}\cite{liang2021pruning}\cite{gupta2022compression}. Pruning involves the elimination of redundant or non-contributory model parameters \cite{reed1993pruning}\cite{liang2021pruning}\cite{gupta2022compression}\cite{li2023model}. Knowledge distillation is a technique where a smaller model is trained to emulate the behavior of a larger, pre-trained model \cite{gou2021knowledge}\cite{alkhulaifi2021knowledge}. Low-rank factorization, on the other hand, approximates the model parameters with lower-rank matrices to compress the model \cite{gupta2022compression}\cite{abadade2023comprehensive}. However, applying these compression techniques in TinyML poses several challenges and limitations. The main challenge is maintaining model robustness and performance while significantly reducing the model’s size and computational requirements \cite{gui2019model}\cite{kwon2021improving}\cite{du2021compressed}. There is a delicate balance between model compression, accuracy, and reliability preservation \cite{ye2019adversarial}\cite{joseph2020going}. Therefore, achieving an optimal trade-off between model size, computational efficiency, and performance robustness and accuracy remains a pivotal challenge in the domain of TinyML. 

To bridge the aforementioned knowledge gaps, this paper employs the principle of physics-based regularization. Prior literature has documented the incorporation of fundamental relationships and characteristics of specific physical systems into the training phase of models to improve their reliability and efficacy \cite{willard2020integrating}\cite{wang2021physics}. Incorporating physics-based relationships into the training process can be achieved by adding them as penalty terms to the objective function \cite{nabian2020physics}\cite{raymond2021applying}\cite{davini2021using}. These terms typically represent the deviation of model predictions from expected behavior based on physical laws or principles. Minimizing this augmented loss function encourages the model to learn representations consistent with the training data and the incorporated physical knowledge \cite{raymond2021applying}\cite{cai2021physics}\cite{cuomo2022scientific}. However, such concepts are yet to be explored within the backdrop of TinyML.

In this context, this work implements a physics-aware TinyML model for predicting ground magnetic perturbations. Furthermore, the authors of this paper have formulated a model compression strategy to prune model elements by leveraging the recognized physical relationships of the investigated system. The accuracy and robustness of the implemented framework were evaluated against a conventional counterpart, and the comparative empirical results have been presented in the subsequent sections. The remainder of this manuscript is structured in the following manner: Section II provides a comprehensive overview of relevant concepts associated with space weather forecasting, the TinyML pipeline, and physics-oriented ML methodologies. Section III describes the methodology adopted in this study, detailing the data acquisition and processing procedures, the formulation of the models and the compression strategy, and their deployment on hardware. Subsequently, Section IV showcases and discusses the empirical findings about the performance of the developed models, also shedding light on potential avenues for future research. The paper concludes in Section V, summarizing the key insights and contributions.

\section{Background and Related Work}

Space weather is the changing environmental conditions in the Earth's magnetosphere (the region of space dominated by the Earth's magnetic field) due to its interaction with the solar wind (a constant stream of charged particles, primarily electrons and protons, emitted by the Sun) \cite{boteler1998effects}\cite{camporeale_machine_2018}\cite{geosciences12010027}. This interaction can lead to disturbances in the Earth's magnetic field and is referred to as geomagnetic disturbances or storms (GMDs) \cite{geosciences12010027}\cite{oliveira_geomagnetically_2017}\cite{pirjola_geomagnetically_2000}. A significant outcome of geomagnetic disturbances (GMDs) manifests as geomagnetically induced currents (GICs), which represent currents induced within conductive structures at the Earth's surface \cite{Wang2020}\cite{jonas}\cite{krausmann2016space}. GICs have significant negative implications for the performance of various technological systems, such as satellite-based operations, navigation systems, and power grids \cite{pirjola_geomagnetically-induced_1989}\cite{pirjola_geomagnetically_2000}\cite{pirjola_effects_2005}. The most prominent event since the dawn of the space age is the geomagnetic storm of 13 March 1989 that caused an electrical power blackout throughout the province of Quebec, Canada \cite{boteler2018dealing}\cite{love2022mapping}\cite{Wang2020}\cite{jonas}\cite{krausmann2016space}. A recent study quantified the global economic impact of a future 1989 Quebec-like event, and the loss would range from \$2.4–\$3.4 trillion over a year \cite{eastwood2017economic}. Therefore, with the increasing technological dependency of our society, predicting GIC occurrence is essential for preventing widespread damage.

Several monitoring and forecasting measures have been adopted to mitigate the risks of the space weather phenomena described above. Common monitoring steps include satellites in the Lagrange points of the Earth's orbit equipped with specialized sensors to collect solar wind data, which are then transmitted to the Earth's ground stations \cite{lyon2000solar}\cite{papitashvili2014omni}. In addition, ground magnetometers are deployed on the Earth's land surface, which measures the strength and direction of the ground magnetic perturbations \cite{engebretson2017future}\cite{yu2008validation}\cite{gjerloev_supermag_2012}. However, obtaining exact measurements of GIC data is challenging, as it requires knowledge and access to the electric power network, which tends to be proprietary information of private energy companies \cite{Siddique2022}\cite{Pinto}. Therefore, different magnetic indices have been used throughout the literature as a proxy measure for GIC \cite{Wintoft2015}\cite{bailey2022}\cite{Keesee}\cite{Siddique2022}\cite{Pinto}. One particular measure is the rate of change in ground horizontal magnetic component ($dB_{H}/dt$), as it has been observed to have a strong correlation with GIC \cite{bailey2022}\cite{Keesee}\cite{Siddique2022}\cite{Pinto}.  

Traditional space weather forecasts rely on physics-based simulation models to interpret the measurements and predict changes in the magnetic field \cite{morley2020challenges}\cite{feng2020current}\cite{LUNDSTEDT20052516}\cite{dredger2023}. Such simulation models are based on first principles and grounded on the fundamental laws of physics, providing a theoretically sound basis for predictions \cite{LUNDSTEDT20052516}. Outputs from the model can often be directly related to physical processes, making it easier to understand and interpret the results \cite{LUNDSTEDT20052516}. In addition, they can be used to predict behaviors in conditions where data might not exist, as they are grounded in theory rather than data alone \cite{morley2020challenges}\cite{feng2020current}\cite{LUNDSTEDT20052516}. Nonetheless, there are limitations to the above approach. The accuracy of the predictions is contingent on the completeness of the intricate underlying physics dynamics \cite{morley2020challenges}\cite{feng2020current}\cite{LUNDSTEDT20052516}. For example, physics-based simulation models require correct initial and boundary conditions derived from observational data. Observational data tend to be noisy or uncertain, and assimilating the data in real-time is challenging, impacting the model's forecasting performance \cite{de2000adaptive}\cite{LUNDSTEDT20052516}. In addition, high-fidelity physics simulations can be computationally intensive and resource-expensive \cite{de2000adaptive}\cite{LUNDSTEDT20052516}\cite{morley2020challenges}.

In recent years, with the advent of Big Data, there has been a rising interest in exploring data-driven approaches, where machine learning (ML) and its variant deep learning (DL) have emerged as a preferred methodology \cite{camporeale_machine_2018}\cite{morley2020challenges}\cite{Keesee}\cite{Pinto}\cite{Siddique2022}. Numerous methodologies have been utilized in forecasting geomagnetic indices and assessing solar wind parameters \cite{camporeale_machine_2018}\cite{morley2020challenges}\cite{Keesee}\cite{Pinto}\cite{Siddique2022}\cite{geosciences12010027}. The efforts included DL-based architectures like artificial neural networks (ANNs) and classical ML regression models like support vector machines (SVMs) and decision trees \cite{camporeale_machine_2018}\cite{camporeale_challenge_2019}. These studies have demonstrated that ML models can provide prediction accuracy comparable to the traditional physics-based simulation models, with much lower resource cost \cite{camporeale_machine_2018}\cite{camporeale_challenge_2019}. ML can detect patterns and relationships in large datasets that might be missed by human analysis or still need to be understood from a first principles perspective \cite{geosciences12010027}. They can be trained on new data to refine and improve predictions. Once trained, many machine learning models can produce results quickly, making them suitable for real-time applications \cite{camporeale_challenge_2019}\cite{geosciences12010027}. Specifically, the past literature comprises work on ML-enabled magnetometer systems for space weather prediction that leverages a form of embedded ML known as Tiny Machine Learning (TinyML). The experimental hardware was configured with a magneto-inductive sensor interfaced with a microcontroller unit (MCU) capable of executing TinyML-based models. The setup was calibrated to conduct instantaneous baseline adjustments of terrestrial magnetic field readings. Subsequent to this correction process, the refined data were employed to forecast $dB_{H}/dt$, serving as a surrogate indicator for GIC levels. The performance of two distinct TinyML models, implemented within this framework, was rigorously evaluated across both real-time and offline forecasting scenarios. In addition, validation of the predicted $dB_{H}/dt$ peak values was carried out employing binary event analysis \cite{Siddique_Sensors}.

TinyML is principally concerned with developing ML models for edge devices with limited resources, such as microcontroller units (MCU) \cite{dutta2021tinyml}\cite{abadade2023comprehensive}\cite{ray2022review}. The TinyML pipeline commences with the model development and training phase, where a machine learning model is conceptualized, designed, and rigorously trained on extensive datasets \cite{dutta2021tinyml}\cite{abadade2023comprehensive}\cite{ray2022review}. This initial phase is computationally demanding and typically conducted on servers or cloud-based platforms with abundant computational resources and memory \cite{abadade2023comprehensive}. Upon training, the model undergoes stringent evaluation using accuracy metrics to ascertain its performance and reliability on unseen data \cite{ray2022review}. In instances of unsatisfactory performance, the model may be refined and retrained until it achieves the desired benchmarks. The model undergoes an optimization and compression process to enable deployment on resource-constrained edge devices like microcontroller units (MCUs). This process primarily includes techniques like quantization, pruning, knowledge distillation, and low-rank factorization, where each method can be applied individually or in combination with one another \cite{dutta2021tinyml}\cite{banbury2020benchmarking}. Quantization refers to constraining or reducing the number of possible values a parameter can take \cite{cheng2017survey}\cite{liang2021pruning}\cite{gupta2022compression}. In relation to neural networks, this usually involves decreasing the accuracy of the weights, for example, shifting from a higher bit of floating-point representations to a lower bit of integer values. This reduces the model's memory footprint, accelerating its inference \cite{cheng2017survey}\cite{liang2021pruning}\cite{gupta2022compression}. Pruning, another optimization method, involves eliminating parts of the neural network, like specific weights or entire neurons, based on criteria like weight magnitude \cite{reed1993pruning}\cite{liang2021pruning}\cite{gupta2022compression}\cite{li2023model}. The goal is to produce a sparser network that retains much of its predictive power. In the third technique, knowledge distillation, a smaller student model is trained to replicate the behavior of a larger, pre-trained teacher model, allowing the former to achieve comparable performance with fewer parameters \cite{gou2021knowledge}\cite{alkhulaifi2021knowledge}. Lastly, low-rank factorization approximates the weight matrices in the network using matrices of a lower rank, effectively capturing critical data features while omitting less crucial parts, thereby reducing computational complexity and storage requirements \cite{gupta2022compression}\cite{abadade2023comprehensive}. Once optimized and compressed, the model is deployed to the target edge device, requiring adaptations to meet the device's limited computational resources, memory, and power \cite{dutta2021tinyml}. Upon deployment, the model is poised for real-time data processing and prediction, capable of analyzing incoming data and making inferences. This enables the edge device to operate remotely and reduces the need for continuous communication with a central server \cite{dutta2021tinyml}.

However, ML models can become overly optimized for training data without careful validation and perform poorly on new, unseen data \cite{geosciences12010027}. Their prediction accuracy depends on the data they are trained on, and biased or unrepresentative data can lead to skewed results \cite{geosciences12010027}. While physics-based simulation models are designed to generalize based on theoretical underpinnings, ML models might not generalize well to conditions significantly different from their training data. This is particularly true for TinyML. The TinyML pipeline, while promising, presents several challenges primarily related to balancing model robustness with resource constraints  \cite{gui2019model}\cite{kwon2021improving}\cite{du2021compressed}. Ensuring model accuracy and reliability in dynamic environments while operating within the limitations of edge devices remains a pivotal challenge and an active area of interest in TinyML \cite{gupta2022compression}\cite{kwon2021improving}\cite{du2021compressed}.

In traditional ML, the integration of physics-based regularization during the training phase has been utilized to enhance the model’s reliability and efficacy by embedding known physical relationships and characteristics of the system into the model \cite{willard2020integrating}\cite{wang2021physics}. These relationships serve as penalty terms to the loss function, guiding the model to learn representations consistent with the training data and the incorporated physical knowledge \cite{raymond2021applying}\cite{cai2021physics}\cite{cuomo2022scientific}. Such methodologies have been successfully applied for forecasting complex natural systems phenomena \cite{jia2021}. For example, the concept has been applied to predict lake temperature by constraining the model with physics-based depth-density relationships \cite{jia2021}. An improved iteration of the same work further integrated temporal physical principles, prioritizing energy conservation in lake thermal dynamics \cite{willard2022}. Additionally, research has been performed where the loss function was amended to address Partial Differential Equations (PDEs) in dynamic system modeling \cite{GENEVA2020109056}. In addition, it has been observed in the past literature that integrating qualitative mathematical properties improves model predictability and generalizability for dynamics such as fluid flow \cite{willard2020integrating}. However, the concept of physics-based regularization is yet to be explored within the TinyML ecosystem. Therefore, this paper has leveraged known physical relationships about space physics to develop a physics-guided TinyML framework. The study utilizes physics-based regularization in the model training phase and also develops a sensitivity-based pruning scheme that leverages the underlying physics-based relationship. The developed framework will aid in advancing the reliability of ML-enabled magnetometer systems for real-time space weather prediction.

\section{Methodology}\label{methodology}

\begin{figure}
\includegraphics[width=8.5cm,height=9.5cm]{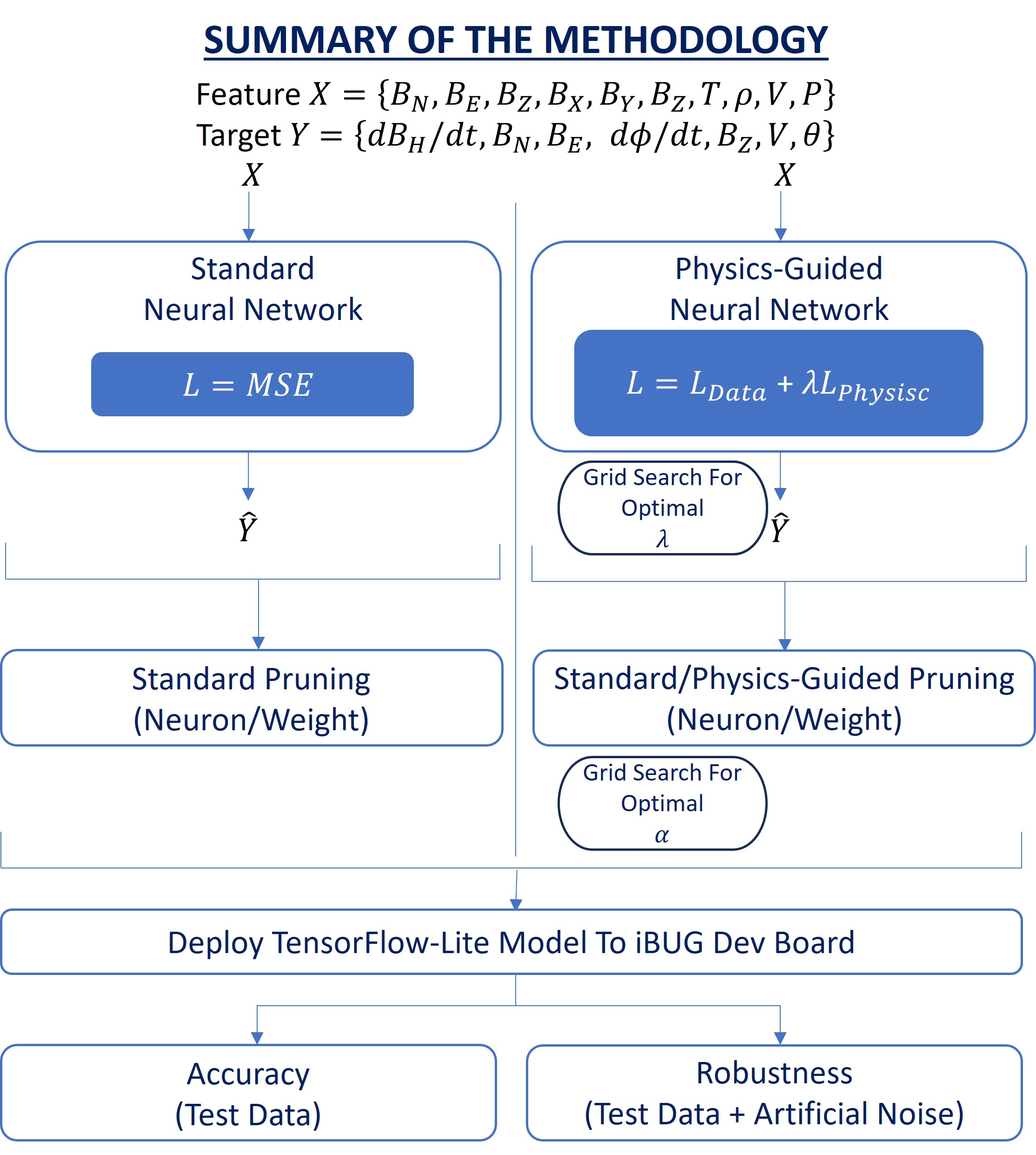}
\caption{Summary of the methodology elaborated in Section-\ref{methodology}. A physics-guided offline model integrates loss functions from dataset accuracy and known physics-based relationships. The model element (neuron or weight) pruning is conducted through traditional sensitivity and constrained violation scores based on the physics-guided regularization term in the loss function. The implemented framework can be utilized to advance the development of machine learning-enabled magnetometer systems.}\label{method_summary}
\end{figure}
\vspace{0.01cm}

\subsection{Data Overview, Acquisition and Processing}

This investigation employs historical terrestrial magnetometer and satellite-derived solar wind measurement data in line with previous research \cite{Keesee}\cite{Pinto}\cite{Siddique2022}\cite{geosciences12010027}. The ground horizontal magnetic component values $dB_{H}/dt$ were derived using the vector components describing the Earth's magnetic field. Generally, the Earth's magnetic field is defined in the Cartesian coordinate system using three vectors: $B_{N}$, $B_{E}$, and $B_{Z}$. The subscripts $N$, $E$, and $Z$ denote the directions in which the values of the components are positive. In a local magnetic reference frame, $N$ is "northwards", $E$ is "eastwards", and $Z$ is "vertically downwards" \cite{}. As shown in Eq-\ref{dbh_dt}, $dB_{H}/dt$ is the resultant of the time derivatives of the two horizontal vectors $B_{N}$, and $B_{E}$, where $dt$ represents a cadence of 1 min \cite{Siddique2022}. The ground magnetometer data were obtained from SuperMAG. SuperMAG is an international alliance of academic and governmental organizations overseeing approximately 600 terrestrial magnetometers globally. This system provides a centralized portal for authenticated Earth's magnetic field variations. The SuperMAG data pipeline comprises a standardized local magnetic coordinate system, uniform time sampling, and a consistent methodology for baseline adjustment \cite{gjerloev_supermag_2012}. For the extraction of the $dB_{H}/dt$ values, as well as for the training and validation of the model, this study leverages the baseline-adjusted magnetic field components from the Ottawa (OTT) magnetometer station, spanning the years 2001 to 2018. The magnetometer at this location operates at a magnetic latitude of 54.98° N. It follows a Universal Time (UT) offset of -5 hours, which implies that the local midnight is synchronized with 05:00 UT \cite{gjerloev_supermag_2012}. The dataset obtained from the OTT station exhibited less than 1\% missing values, which have been rectified through linear interpolation techniques.

\begin{equation}\label{dbh_dt}
\frac{dB_{H}}{dt} = \sqrt{(\frac{dB_{N}}{dt})^2 + (\frac{dB_{E}}{dt})^2}
\end{equation}

The solar wind data comprises the interplanetary magnetic Field (IMF), plasma measurements, and derived parameters. The IMF extends the Sun's magnetic field, propagated through space via the solar wind \cite{Keesee}\cite{Pinto}. It significantly influences Earth's magnetosphere and can affect geomagnetic activities. Its orientation and strength are critical factors in triggering geomagnetic storms \cite{Keesee}. Like the Earth's magnetic field, the IMF is a vector quantity. It is commonly analyzed in specific coordinate systems like geocentric solar magnetospheric (GSM) or Geocentric Solar Ecliptic (GSE). Both GSM and GSE utilize a Cartesian framework, where the IMF is defined using three components: $B_{X}$, $B_{Y}$, and $B_{Z}$ \cite{hapgood1992space}\cite{NASA}. In both frames, the $X$-axis is aligned from Earth towards the Sun. The GSE coordinate system is ecliptic-aligned, with its $Y$-axis parallel to the ecliptic plane and oriented towards dusk. Its $Z$-axis runs parallel to the ecliptic pole \cite{hapgood1992space}. Contrastingly, in the GSM system, the $Y$-axis is set to be orthogonal to Earth's magnetic dipole, while its positive $Z$-axis aligns with the direction of the northern magnetic pole \cite{hapgood1992space}. These coordinate systems can be transformed through suitable rotation about their respective $X$-axes \cite{hapgood1992space}. For this paper, which centers on magnetospheric disturbances, the authors opted for the IMF data represented in the GSM coordinate framework. As for the plasma and derived parameters, proton temperature ($T$), density ($\rho$), flow speed ($V$), and pressure ($P$) were considered. The solar wind data were collected from the OMNIWeb repository, managed by NASA's Space Physics Data Facility. The repository curates near-Earth solar wind magnetic field and plasma parameter data from several spacecraft located in geocentric orbits \cite{NASA}\cite{King2005}. Similar to the ground magnetometer data, solar wind data of 1 min cadence from the years 2001-2018 were employed for this study. The IMF and plasma data have approximately 8\% and 20\% of missing data, which were addressed through linear interpolation. The compiled dataset used in this paper, $D$, comprised the following variables as set elements $\{B_{N},B_{E},B_{Z},B_{X},B_{Y},B_{Z},T,\rho,V,P, dB_{H}/dt\}$. 

\subsection{Neural Networks, Physics-Guided Regularization, and Newell Coupling Function: Overview and Implementation}\label{B}

In conventional neural network (NN) architectures, the network consists of an input and output layer, interconnected by one or more hidden layers \cite{siddique2021classification}\cite{geosciences12010027}. Each layer comprises multiple continuous variables called neurons \cite{siddique2021classification}\cite{geosciences12010027}. The number of neurons in the input layer corresponds to the dimensionality of the features in the variable $X$, and the output layer's size is determined by that of the target variable $Y$, both of which are components of the dataset $D$ \cite{geosciences12010027}. The strength of the connection between neurons of consecutive layers is represented using a set of continuous model parameters or weights $W$ \cite{geosciences12010027}. The NN aims to attain the optimal $W$, which in turn will express the most accurate relationship between $Y$ and $X$ \cite{geosciences12010027}. During the training phase, the NN undergoes forward and backward propagation procedures. Initially, an arbitrary weight set $W$ is established, and the feature vector $X$ from the dataset is fed into the network's input layer. Subsequent hidden layers employ an activation function $f$, which transforms the linear combination of the input $X$ and weights $W$, augmented by the bias term $b$, into a resulting hidden state $H$ (as defined in Eq-\ref{Z} and \ref{H}) \cite{geosciences12010027}. Activation functions such as $tanh$, $sigmoid$, and the $Rectified Linear Unit (ReLU)$ are typically utilized to capture the inherent non-linearities between the feature and target variables $X$ and $Y$ \cite{siddique2021classification}.

\begin{equation}\label{Z}
Z = W \cdot X + b
\end{equation}

\begin{equation}\label{H}
H = f(Z)
\end{equation}

The forward propagation concludes when the output layer generates an estimated target variable $Y$. This predicted output is then compared against its corresponding validation set value $Y'$ by employing a loss function $\mathcal{L}$ \cite{}. A commonly used loss function for this purpose is the mean-squared error (MSE). The primary aim is to update the weight set $W$ during backward propagation in such a manner that the loss function $\mathcal{L}$ is minimized (refer to Eq-\ref{mse} for its mathematical formulation) \cite{geosciences12010027}. The minimization of the loss function is executed through a numerical optimization technique known as gradient descent \cite{geosciences12010027}. In this context, the loss $\mathcal{L}$ is regarded as a function of a specific weight $w$, and the goal is to attain the weight value that minimizes this loss function. Given an initial weight $w_{initial}$ and a subsequent weight $w_{next}$ progressing toward the minimum point of the loss function, the gradient descent update step is articulated in Eq-\ref{grad_desc} \cite{geosciences12010027}. In this equation, the learning rate $\epsilon$ is a hyperparameter that governs the rate at which the model adapts its weight parameters. This iteration of forward and backward propagation is conducted over multiple epochs, where each epoch signifies a complete pass through the training dataset \cite{brownlee2018difference}. The iterative process continues until the value of the loss function reaches a predetermined threshold.

\begin{equation}\label{mse}
\mathcal{L}(W) = \frac{1}{N}\sum_{i=1}^{N} (x_{i}' - g(f(x_{i},w_{f}),w_{g}))^2
        = \frac{1}{N}\sum_{i=1}^{N} (x_{i}' - \hat{x}_{i}')^2
\end{equation}

\begin{equation}\label{grad_desc}
w_{next} = w_{initial} - \epsilon\nabla_{w_{inital}}\mathcal{L}(w_{inital}).
\end{equation}

While traditional neural networks (NNs) are primarily designed to fit data, they often lack consideration for the underlying physical properties of the system, thereby lacking robustness \cite{karniadakis2021physics}. Physics-guided regularization can be leveraged to incorporate domain knowledge within the model learning process \cite{raymond2021applying}\cite{cai2021physics}\cite{cuomo2022scientific}. Regularization as a technique is utilized in ML to mitigate the risk of overfitting \cite{tian2022comprehensive}. Overfitting is a prevalent challenge, particularly in high-dimensional settings, where models, while minimizing the training error, inadvertently fit the noise inherent in the training data, compromising their generalization capability \cite{tian2022comprehensive}. Regularization techniques penalize model complexity, steering models towards simplicity and enhanced generalization. A physics-guided regularization aims to constrain the learned model with known physical relationships, where the relationships are typically via an established equation. Structurally and procedurally, the NN remains the same in its architecture and iterative training approach. The distinction lies in formulating the loss function $\mathcal{L}$. The loss function becomes a composite of two terms:

\begin{itemize}
\item Data Term ($\mathcal{L}_{Data}$): It quantifies the degree to which the NN approximates its parameters based on the available data points \cite{karniadakis2021physics}. It is commonly expressed through the mean-squared error (MSE) for regression, as detailed in Eq-\ref{mse}.
\item Physics Term ($\mathcal{L}_{Physics}$): It constrains the NN to the governing mathematical relationship describing the underlying domain principles. This is achieved by formulating the residual between the equation's left and right-hand side term(s) \cite{karniadakis2021physics}.
\end{itemize}

Thus, the optimization ensures that the trained model fits the data and conforms to known physical relationships. Mathematically, the updated loss function with the physics-guided regularization $\mathcal{L}$ can be expressed as shown in Eq-\ref{phy_reg}. Here, $\lambda$ ($0 \leqslant\lambda\leqslant 1$) is a weighting parameter that balances the importance between the data and physics terms. A $\lambda$ value of $0$ converts the model into a traditional NN. In contrast, a value of $1$ signifies that the $\mathcal{L}_{Data}$ and $\mathcal{L}_{Physics}$ have the same degree of importance during model training. The value of $\lambda$ can be determined via hyper-parameter tuning.

\begin{equation}\label{phy_reg}
\mathcal{L} = \mathcal{L}_{Data} + \lambda\mathcal{L}_{Physics}
\end{equation}

Considering that the primary objective of this study is to forecast terrestrial magnetic disturbances induced by space weather events, the Newell Coupling Function serves as a suitable mathematical formulation for modeling the physical system under investigation \cite{newell2007nearly}. The function serves as a quantitative framework for assessing the rate at which energy is transferred from the solar wind into Earth's magnetosphere \cite{newell2007nearly}\cite{spencer2009evaluation}. This energy coupling is a critical mechanism underlying various space weather phenomena, including geomagnetic storms and auroral activities. The function estimates how efficiently the solar wind couples with the magnetosphere \cite{newell2007nearly}\cite{spencer2009evaluation}. The function is defined as exhibited in Eq-\ref{newell_funct} \cite{newell2007nearly}\cite{spencer2009evaluation}. In Eq-\ref{newell_funct}, $\frac{d\Phi}{dt}$ represents the rate of magnetic reconnection at the magnetopause. The solar wind flow speed $V$, and $V^{\frac{4}{3}}$ emphasizes that a faster wind transfers energy more efficiently. The GSM coordinate system represents the $Z$-component of the IMF $B_{Z}$. The clock angle $\theta$ essentially describes the tilt of the IMF as it approaches the Earth, calculated as shown in Eq-\ref{clock_angle}, and $sin^{\frac{8}{3}}(\frac{\theta}{2})$ accounts for the orientation of the IMF, with the maximum coupling occurring when the IMF is directly southward.

\begin{equation}\label{newell_funct}
\frac{d\Phi}{dt} = V^{\frac{4}{3}}B_{z}^{\frac{2}{3}}sin^{\frac{8}{3}}(\frac{\theta}{2})
\end{equation}

\begin{equation}\label{clock_angle}
\theta = \tan^{-1}(\frac{B_{Y}}{B_{Z}})
\end{equation}

In this research, the authors implemented a standard Neural Network (NN) and a Physics-Guided Neural Network (PGNN), each maintaining identical architectural configurations. The features $X$ and the targets $Y$ were extracted from the dataset $D$. The feature set $X$ was multivariable, and incorporated ground magnetometer components ($B_{N},B_{E},B_{Z}$), Interplanetary Magnetic Field (IMF) components ($B_{X},B_{Y},B_{Z}$), along with solar wind parameters ($T,\rho,V,P$). Conversely, the target set $Y$ was multivariate and comprised of the rate of horizontal ground magnetic perturbation ($dB_{H}/dt$), the terrestrial magnetic components in the North and East ($B_{N}, B_{E}$), the rate of magnetic reconnection at the magnetopause ($d\Phi/dt$), the IMF component ($B_{Z}$), the solar wind speed ($V$), and the clock angle ($\theta$). Notably, the data corresponding to $d\Phi/dt$ were derived utilizing Eq-\ref{newell_funct} and \ref{clock_angle}. Consequently, each of the NNs was configured with an input layer incorporating ten neurons and an output layer with seven neurons. The models consisted of 3 hidden layers, consisting of 30 neurons each. The hidden layers used ReLU as its activation function. The conventional NN was developed to use MSE as its loss function, and the Adam optimizer was used for the minimization process. In contrast, a composite loss function $\mathcal{L}_{PGNN}$ was developed for the PGNN, similar to the one illustrated in Eq-\ref{phy_reg}. The $\mathcal{L}_{Data}$ in Eq-\ref{phy_reg} is the sum of the individual MSE of each of the target variables and their model forecasted counterpart. In Eq-\ref{l_data}, the mathematical formulation for the implemented $\mathcal{L}_{Data}$ have been presented, where $Y_{i,j}$ and $\hat{Y}_{i,j}$ are the observed and predicted $i^{th}$ data point respectively, the $j^{th}$ output variable. In the equation, $i = \{1, 2,...N\}$ ($|i|= N$), and $j = \{dB_{H}/dt, B_{N}, B_{E}, d\Phi/dt, V, B_{Z},\theta\}$ ($|j|= K$). In developing the $\mathcal{L}_{Physics}$, for the Physics-Guided Neural Network (PGNN), the authors prioritized the incorporation of domain knowledge pertinent to solar wind-magnetospheric interactions and their consequential impacts on ground magnetic perturbations. Accordingly, Eq-\ref{dbh_dt} and \ref{newell_funct} were leveraged to formulate two regularization terms, constraining the model to adhere to established physical relationships (See Eq-\ref{l_physics}). The formulation of the two regularization terms $R_{1}$ and $R_{2}$ which combines to form the $\mathcal{L}_{Physics}$, are shown in Eq-\ref{reg_1} and \ref{reg_2}. The constraint $R_{1}$ ensures that the forecasted rate of change of the horizontal component of the ground magnetic field, $dB_{H}/dt$, is consistent with the predicted ground magnetic component. On the other hand, $R_{2}$ guarantees that the projected interaction between the solar wind and magnetosphere, in conjunction with the forecasted Interplanetary Magnetic Field (IMF) and solar wind parameters, adheres to Newell's coupling function. During training, these regularization terms penalize the model for significant deviations from established physics-based relationships and characteristics.

\begin{equation}\label{l_data}
    \mathcal{L}_{Data} = \frac{1}{K}\sum_{j}^{K}\frac{1}{N}\sum_{i}^{N}(Y_{i,j}-\hat{Y}_{i,j})^2
\end{equation}

\begin{equation}\label{l_physics}
    \mathcal{L}_{Physics} = R_{1} + R_{2}
\end{equation}

\begin{equation}\label{reg_1}
    R_{1} = ||(\frac{d\hat{B}_{H}}{dt})^2 - (\frac{d\hat{B}_{N}}{dt})^2 - (\frac{d\hat{B}_{E}}{dt})^2||
\end{equation}

\begin{equation}\label{reg_2}
    R_{2} = ||\frac{d\hat{\Phi}}{dt} - \hat{V}^{\frac{4}{3}}\hat{B}_{Z}^{\frac{2}{3}}\sin^{\frac{8}{3}}(\frac{\hat{\theta}}{2})||
\end{equation}

For model training, 80\% of the data were leveraged, and the remainder were used for testing. For validation during model training, 20\% of the training set was used. The train and test set elements underwent distinct normalization steps to avoid data leakage. The applied normalization, specifically a min-max normalization, is represented by Eq.-\ref{minmax}. In this equation, $x$ symbolizes a data point member of a specific variable $X$, while $x_{norm}$ denotes its normalized equivalent. Additionally, $x_{max}$ and $x_{min}$ represent the maximum and minimum values within the set, respectively. The weighting parameter $\lambda$ in $\mathcal{L}_{PGNN}$ was tuned using grid search. Grid search is an optimization approach that evaluates predefined hyperparameters to ascertain the optimal combination for a specific model and dataset. By rigorously testing all potential hyperparameter combinations, this technique ensures comprehensive exploration. Typically, it is paired with cross-validation to validate the consistency of model performance across varied data partitions. The procedure was executed utilizing the GridSearchCV function from the Scikit-learn library in Python. Considering that the PGNN model was developed in the TensorFlow framework, the KerasRegressor function was employed to ensure compatibility with Scikit-learn functionalities.  
 
\begin{equation}\label{minmax}
    x_{norm} = \frac{x - x_{min}}{x_{max} - x_{min}}
\end{equation}

\subsection{Physics-Guided Model Compression and Tiny Machine Learning (TinyML): Network Pruning}

Model compression encompasses a spectrum of methodologies aimed at diminishing the memory footprint and computational demands of Neural Networks (NNs), thereby accelerating inference and facilitating deployment on edge devices \cite{gupta2022compression}. A prevailing strategy within this domain is network pruning. Network pruning focuses on eliminating model elements distinguished as model weights or neurons that exhibit a negligible influence on the accuracy of model predictions \cite{zhu2017prune}. The academic literature delineates a variety of approaches to model pruning. These approaches can be broadly classified based on three distinctive criteria: 1) the structure of the pruned network, categorized as either symmetric or asymmetric; 2) the model element being pruned, either a neuron or a weight; and 3) the timing of the pruning process, distinguished as offline (static) or online (dynamic) pruning. It is important to note that there exists a degree of overlap among these categories \cite{liang2021pruning}. After these primary categorizations, the methodologies employed for pruning within any of the categories above can be further classified into either sensitivity-based or penalty-based methods \cite{liang2021pruning}\cite{augasta2013pruning}. 

Sensitivity-based methods operate by initially training the network, estimating the sensitivities, and then excising weights or nodes identified as having minimal impact on the error function \cite{augasta2013pruning}. In essence, this approach modifies an already trained network by assessing which elements, when removed, would incur the slightest alteration to the error function and subsequently pruning them \cite{augasta2013pruning}.

Contrastingly, penalty-term methods introduce modifications to the loss function \cite{augasta2013pruning}. This alteration ensures that the modified function drives the redundant model elements toward zero during the backpropagation process, effectively eliminating them during the training phase. This approach inherently rewards the development of sparse representations and efficient solutions, aligning the network’s optimization process with the goal of model compression \cite{augasta2013pruning}.

For this research, two distinct network pruning strategies were implemented. The first strategy adhered to a conventional sensitivity-based approach, systematically assessing the impact of the elimination of an individual model element category (neurons or weights) on the overall network performance. The subsequent strategy represents a physics-guided hybrid approach developed by the authors of this paper. This methodology amalgamates both sensitivity and penalty-based techniques by leveraging the physics-guided regularization terms in the previous section. The conventional and physics-guided pruning schemes are summarized below.

\subsubsection{Standard Model Element Pruning Scheme}

Post initial training, the algorithm calculates the importance score $S_{ij}$ for each member $j$ under a particular model element $i$ in the neural network (NN). For a specific pruning iteration, $i$ represents either the model neurons or weights. The element importance scores $S_{ij}$ is computed as the absolute value of the partial derivative of the loss function $\mathcal{L}$ with respect to the output $o_{ij}$ of the $j^{th}$ model element, denoted as $S_{ij} = |\frac{\delta\mathcal{L}}{\delta o_{ij}}|$. The elements are then ranked based on their importance scores $S_{ij}$, and a predetermined proportion $r$ of the total elements, specifically those with the lowest scores, are pruned from the neural network. Following the pruning process, the model undergoes further refinement using the training dataset, allowing for the optimization of the weights and the performance levels. In the present research, the implemented standard NN and the physics-guided NN model underwent optimization and compression, utilizing the above element pruning methodology. The pseudocode of the implemented scheme is provided in Algorithm 1.

\begin{algorithm}\label{std_prun}
\caption{Standard Model Element Pruning Scheme}
\begin{algorithmic}[1]
\REQUIRE Training dataset, pruning ratio $r$

\STATE \textbf{Model Training:}
\STATE Define the loss function $L$ (MSE).
\STATE Train the neural network $NN$ to minimize $L$ using the training dataset.

\STATE \textbf{Identify Prunable Model Elements:}
\FOR{each set member $j$ for a particular model element $i$ in the $NN$, where $i$ can either be model neurons or weights}
    \STATE Compute element importance score: $S_{ij} = \left| \frac{\partial L}{\partial o_{ij}} \right|$, where $o_{ij}$ is the output of the $j^th$ member of a specific model element $i$.
\ENDFOR

\STATE \textbf{Prune Model Elements:}
\STATE Rank elements based on $S_{ij}$ in ascending order.
\STATE Determine the number of elements to prune $j_p = r \times \text{total number of elements}$.
\STATE Prune the lowest $j_p$ score elements from $NN$.

\STATE \textbf{Fine-tuning:}
\STATE Fine-tune the pruned model on the training dataset.

\STATE \textbf{Output:}
\STATE Pruned and fine-tuned neural network.
\RETURN
\end{algorithmic}
\end{algorithm}

\subsubsection{Physics-Guided Model Element Pruning Scheme}\label{prun_sec}

The physics-guided model element pruning algorithm incorporates additional physical constraints during model training. The loss function in this approach is augmented by a term that penalizes violations of the incorporated physical relationships, represented in Eq-\ref{phy_reg}. Like the standard algorithm, the physics-guided approach computes an importance score $S_{ij}$ for the members of a particular model element $i$, using the loss function $L$. Additionally, it calculates a constraint violation score $C_{ij} = |\frac{\delta\mathcal{L}_{physics}}{\delta o_{ij}}|$ for each model element, reflecting its contribution to the violation of the physical constraints. A combined score $T_{ij}= S_{ij} + \alpha C_{ij}$ is then computed, which balances the element's importance and its adherence to physical constraints, with $\alpha$ as a balance factor which ranges from $0$ to $1$. The model elements are pruned based on $T_{ij}$, and the pruned network is retrained to fine-tune the remaining weights while respecting data-driven loss and physical constraints. The value of $\alpha$ is determined using a grid search methodology similar to the one described in section-\ref{B}. In the scope of this study, the physics-guided pruning methodology was exclusively employed on the implemented physics-guided NN. The standard neural network was exempted from this pruning approach due to the absence of a physics-based regularization term in its loss function, which is essential for computing $C_{ij}$. The pseudocode for the physics-guided pruning scheme is summarized in Algorithm 2.

\begin{algorithm}\label{phy_prun}
\caption{Physics-Guided Model Element Pruning Scheme}
\begin{algorithmic}[1]
\REQUIRE Training dataset, physical constraints $\mathcal{L}_{physics}$, balance factor $\alpha$, pruning ratio $r$, regularization parameter $\lambda$

\STATE \textbf{Model Training:}
\STATE Define loss function $L = \mathcal{L_{MSE}} + \lambda \mathcal{L}_{physics}$ where $\mathcal{L}_{physics}$ is the penalty for violating physical constraints.
\STATE Train the neural network $NN$ to minimize $L$ using the training dataset.

\STATE \textbf{Identify Prunable Model Elements:}
\FOR{each set member $j$ for a particular model element $i$ in the $NN$, where $i$ can either be model neurons or weights}
    \STATE Compute element importance score: $S_{ij} = \left| \frac{\partial L}{\partial o_{ij}} \right|$
    \STATE Compute constraint violation score: $C_{ij} = \left| \frac{\partial \mathcal{L}_{physics}}{\partial o_{ij}} \right|$
    \STATE Compute combined score: $T_{ij} = S_{ij} + \alpha C_{ij}$
\ENDFOR

\STATE \textbf{Prune Model Elements:}
\STATE Rank elements based on $T_{ij}$ in ascending order.
\STATE Determine the number of elemnts to prune $j_p = r \times \text{total number of elements}$.
\STATE Prune the lowest $j_p$ score neurons from $NN$.

\STATE \textbf{Fine-tuning:}
\STATE Retrain the pruned network to fine-tune weights.

\STATE \textbf{Output:}
\STATE Pruned and fine-tuned neural network.
\RETURN
\end{algorithmic}
\end{algorithm}

\subsection{Hardware and Deployment}

\begin{figure}
\begin{center}
\includegraphics[width=5.5cm,height=4cm]{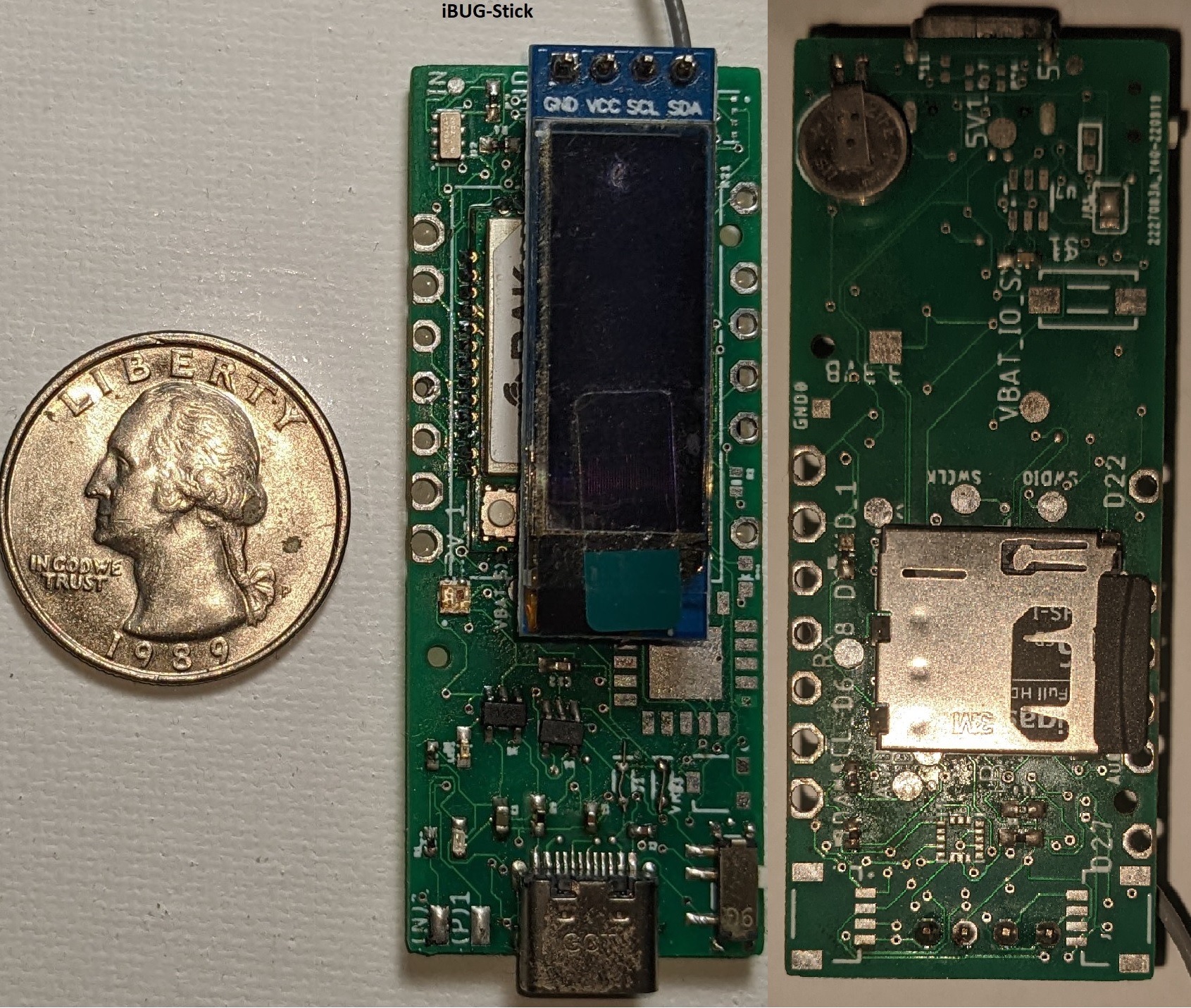}
\end{center}
\caption{The iBUG-Stick development board (Front and Back Views) is presented adjacent to a US cent coin, serving as a visual reference for size comparison.}\label{ibug}
\end{figure}

For evaluation, both offline and TinyML model variants were taken into consideration. Here, "offline" denotes a model trained locally, which neither underwent pruning nor was deployed on an MCU. Three unique TinyML variants of the two implemented model architectures were deployed onto an MCU development board. These model variants comprise: 1) a standard NN, optimized using standard model element pruning; 2) a physics-guided NN, refined with standard model element pruning; and 3) a physics-guided NN, optimized utilizing physics-guided model element pruning. The chosen development board for deployment was the iBUG-stick, as exhibited in Fig-\ref{ibug}. The iBUG board is an ML-capable Internet of Things (IoT) sensing platform with a history of utilization in real-time environmental monitoring, as documented in past research \cite{yousuf_ibug:_2022}. This board has a RAK11300 Long Range Wide Area Network (LoRaWAN) module and a dual-core Raspberry Pi RP2040 MCU operating at 133MHz. For testing the TinyML variants, relevant input from the designated test dataset was streamed to the model through the board's USB serial communication. The robustness of each model iteration was assessed by generating predictions from test data subjected to varying degrees of noise. A summary of the overall methodology implemented in this paper is illustrated in Fig-\ref{method_summary}.

\section{Results and Discussion}

\begin{figure} 
\includegraphics[width=10cm,height=7cm]{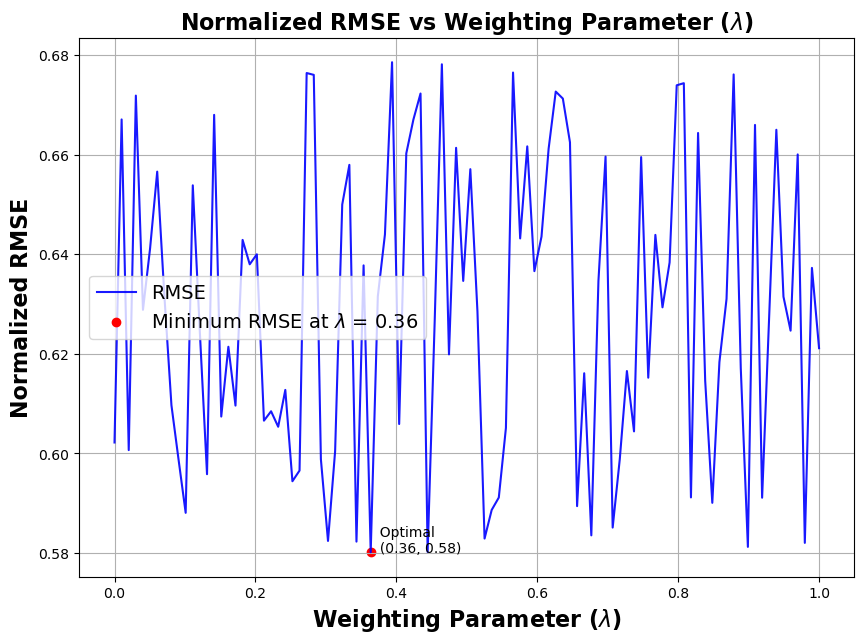}
\caption{Normalized Root Mean Squared Error VS Weighting Parameter ($\lambda$) for the the physics-guided regularization term, $\mathcal{L}_{physics}$, in the loss function of the physics-guided model. The plot shows the optimal $\lambda$ value which has the minimum RMSE.}\label{lambda}
\end{figure}
\vspace{0.01cm}

\begin{figure} 
\includegraphics[width=9cm,height=11.5cm]{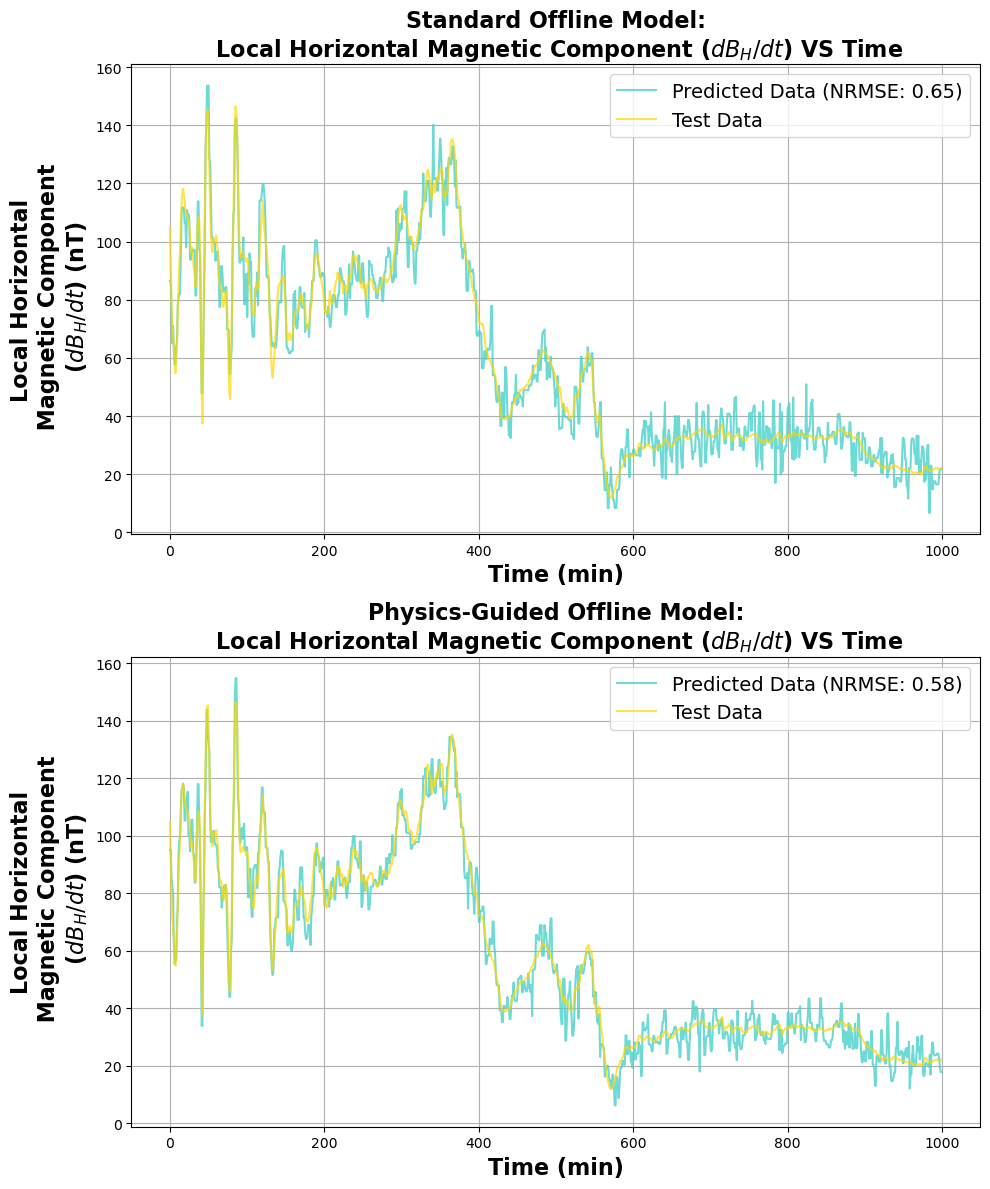}
\caption{Test and Predicted Data of Local Ground Horizontal Magnetic Component ($\frac{dB_{H}}{dt}$) (nT) VS Time (mins). The figure provides a comparison of the performance between the conventional and physics-guided offline models prediction against the ground truth.}\label{offline_dBH_dt}
\end{figure}
\vspace{0.01cm}

\begin{figure} 
\includegraphics[width=9cm,height=11.5cm]{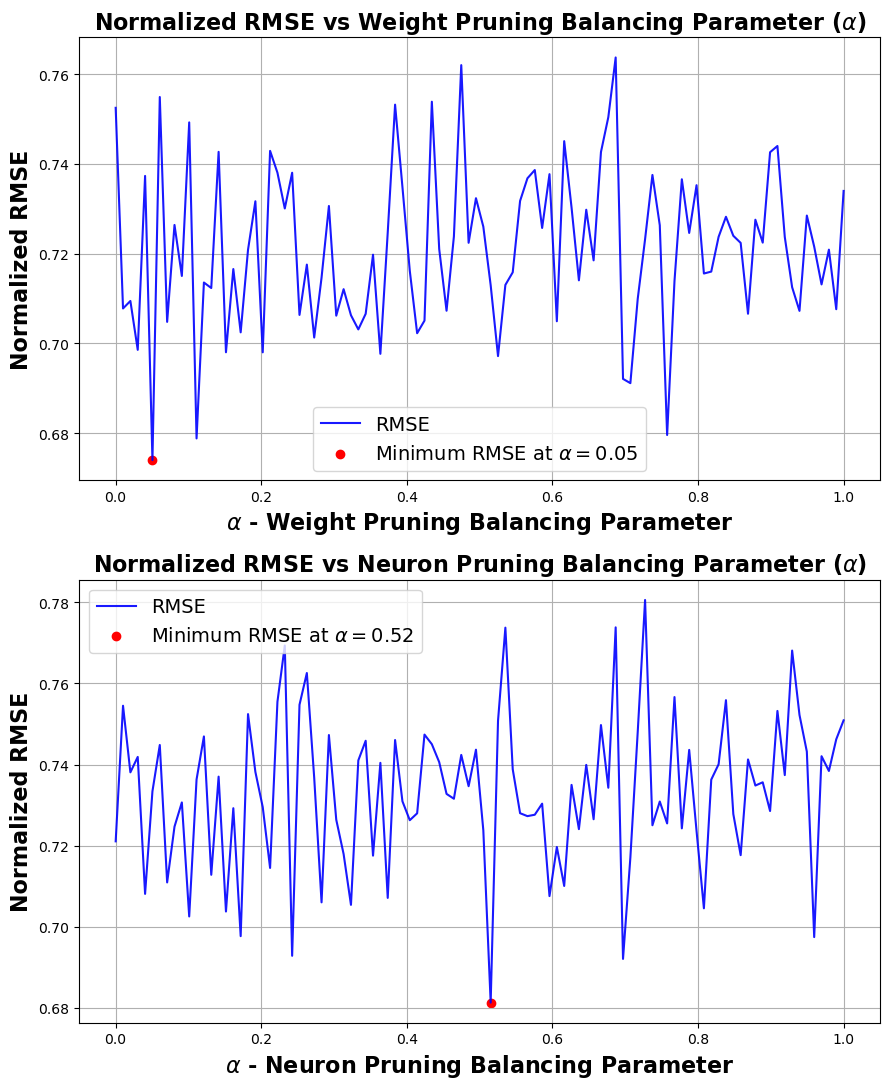}
\caption{Normalized Root Mean Squared Error (Normalized RMSE) VS Model Element Pruning Balancing Parameter ($\alpha$). The plot shows the optimal $\alpha$ value both for the case of weight pruning, and neuron pruning, distinctly applied to the physics-guided offline Neural Network. }\label{alpha}
\end{figure}
\vspace{0.01cm}

\begin{figure} 
\includegraphics[width=9.5cm,height=12cm]{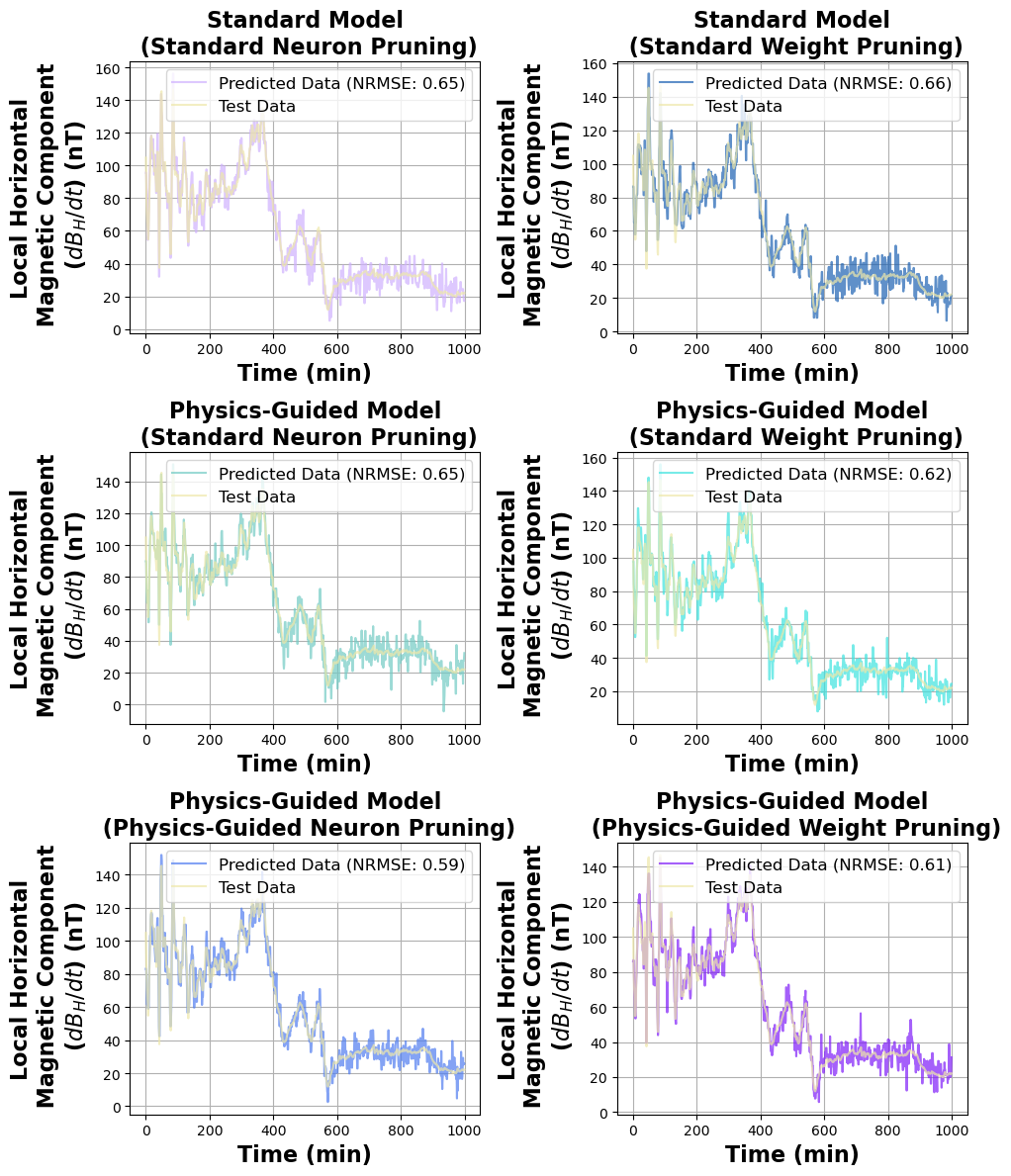}
\caption{Local Ground Horizontal Magnetic Component ($\frac{dB_{H}}{dt}$) (nT) VS Time (mins). Compares the prediction of each model variant under different pruning scheme, against the test data.}\label{pruning_dBH_dt}
\end{figure}
\vspace{0.01cm}

\begin{figure} 
\includegraphics[width=9cm,height=12cm]{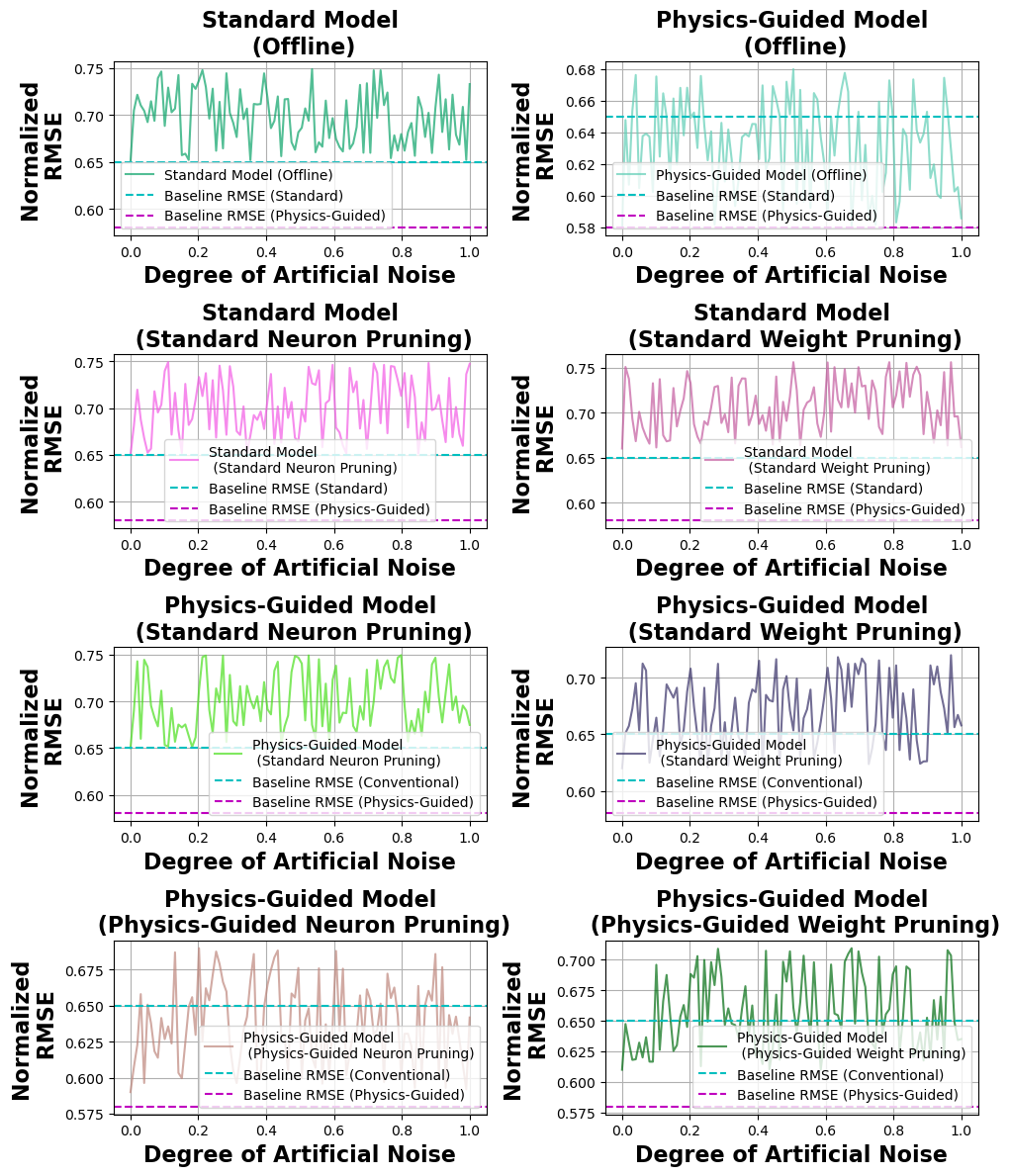}
\caption{Comparison of Robustness To Varying Degree of Artificial Noise In The Test Data}\label{robustness}
\end{figure}
\vspace{0.01cm}

Fig-\ref{lambda} depicts the outcomes of the grid search conducted to determine the optimal value for the hyperparameter $\lambda$. This parameter serves as a weighting coefficient for the physics-guided regularization term, $L_{physics}$, within the framework of the implemented physics-guided NN. The graph expresses the variation in normalized RMSE values for distinct $\lambda$ values spanning the interval $[0, 1]$. Notably, $\lambda = 0.36$ emerges as optimal, yielding the lowest normalized RMSE of $0.58$.

In Fig-\ref{offline_dBH_dt}, using the optimal value of $\lambda$, a comparison of the predicted versus the actual test data for the Local Horizontal Magnetic Component ($\frac{dB_{H}}{dt}$) over time (minutes), have been presented. This figure exhibits results from the two distinct modeling approaches: a standard NN and a physics-guided NN. Upon closer inspection of the standard offline model, the predictions, characterized by a normalized RMSE of $0.65$, generally adhere to the initial variations observed in the $\frac{dB_{H}}{dt}$ values. However, intermittent discrepancies arise, particularly noticeable when the actual $\frac{dB_{H}}{dt}$ values are relatively low. These discrepancies manifest as marked fluctuations in the predicted data. In contrast, the physics-guided model showcases improved conformity with the test data, especially during the early variations in $\frac{dB_{H}}{dt}$. Its lower NRMSE of $0.58$ quantitatively validates this. When considering the segments of the dataset with diminished $\frac{dB_{H}}{dt}$ values, the physics-guided model showcases reduced fluctuations compared to the standard model. These reduced fluctuations are due to the domain-specific knowledge incorporated into the loss function as a regularization term. The observed efficacy in the physics-guided model affirms the utility of integrating domain-specific, physics-based principles into the predictive modeling framework. 

Fig-\ref{alpha} details the outcomes of the grid search aimed to obtain the optimal value of the hyperparameter $\alpha$ for the physics-guided pruning strategy applied to the physics-guided model. As mentioned in Section-\ref{prun_sec}, $\alpha$ serves as a balancing coefficient for the constraint violation scores in the pruning scheme. The visual representation displays results for two intrinsic model elements: model weights and neurons. Both sub-figures plot normalized RMSE values as a function of the $\alpha$ parameter, which spans the interval [0, 1]. An optimal $\alpha$ of $0.05$ emerges for model weight adjustments, while neuron adjustments find their equilibrium at an $\alpha$ value of $0.52$. These findings indicate that the best predictive performance is achieved with a minimal weight adjustment parameter, whereas a mid-range adjustment is preferable for neurons. It can be inferred that selecting a pruning strategy and its associated balancing coefficient can exert substantial influence over the model's predictive prowess. Additionally, determining the most relevant pruning approach and its corresponding hyperparameter is contingent on the model component under consideration and the specificity of the problem domain.

Fig-\ref{pruning_dBH_dt} offers a comprehensive comparison of predicted values for $\frac{dB_{H}}{dt}$ relative to the ground truth, drawing from forecasts produced by the two distinct model architectures subjected to various pruning schemes. As mentioned in Section-\ref{prun_sec}, the standard NN underwent optimization solely through conventional pruning techniques, either neuron or weight-based. In contrast, the physics-guided variant was pruned using both the standard and the physics-guided pruning methodologies. The standard model, post neuron and weight pruning, register normalized RMSE scores of $0.65$ and $0.66$, respectively, indicating a predictive performance on par with their offline counterpart. Their forecasting patterns, irrespective of the pruning technique, align with those generated by the offline standard model. In the case of the physics-guided model, its performance under standard neuron pruning mirrors that of its pruned standard counterparts. However, an improvement in accuracy is observed with the standard weight pruning, evidenced by a RMSE score of $0.62$. However, the best performance is achieved when the model is subjected to physics-guided pruning, both neuron and weight-based, yielding RMSE scores of $0.59$ and $0.61$, respectively. In addition, the physics-guided neuron-pruned model matches the precision of its offline version. This underscores the utility of combining physics-based constraints within the pruning strategy, which ensures that the model's offline efficacy is retained post-optimization and compression.

Fig-\ref{robustness} examines the robustness of each model variant when the test data is subjected to varying degrees of artificial noise. Each subplot displays the normalized RMSE values across a range of artificial noise, from 0 to 1. Each subplot features two baseline RMSEs – one corresponds to the normalized RMSE from the offline standard configuration, and the other represents the physics-guided offline variant derived from evaluations using the unaltered test data. Upon examining the results, several observations emerge. All the models demonstrate a general increase in normalized RMSE when evaluated with artificial noise. This trend underscores the inherent challenge of preserving prediction accuracy in the presence of noise. For the offline configuration, the standard and physics-guided models present oscillatory behaviors in normalized RMSE with increasing noise. However, the physics-guided model consistently maintains a marginally lower normalized RMSE, indicating its higher resilience to noisy inputs. Among the pruning methods, physics-guided neuron pruning and physics-guided weight pruning exhibit the most stability against noise perturbations. In contrast, the standard pruning techniques, especially weight pruning, appear slightly more susceptible to noise-induced deviations. Among all the combinations of model architectures and pruning methods, the physics-guided model consistently demonstrates superior robustness in the face of noisy data when optimized using the physics-guided neuron pruning approach.

The insights derived from the results indicate that the physics-guided TinyML framework presents a promising avenue for developing robust machine learning-enabled magnetometer systems. The framework balances model size and computational needs, showing competitive performance even under constrained resources. The study also revealed the significance of optimizing hyperparameters and pruning strategies to achieve reliable and accurate predictions. Future research endeavors can expand upon the findings of this study by developing a physics-guided TinyML framework for satellite-based deployments for real-time processing and inference of solar wind data. Moreover, integrating more advanced machine learning models and algorithms could help adapt to the diverse and dynamic nature of geomagnetic phenomena, thereby improving the predictive capabilities of magnetometer systems. Exploration of real-world applications and deployment scenarios will also be essential. This includes assessing the framework's performance in various environmental conditions and its adaptability to different magnetometer configurations and specifications. Additionally, further studies could explore the integration of more complex first principles-based physics constraints, which are currently used in simulation-based numerical models for space weather forecasting. Such an endeavor would enrich the model's understanding and representation of the underlying physical processes. Finally, addressing challenges associated with real-time data processing and testing the models developed for edge devices with varying computational resources remains crucial for future work. This would involve developing strategies for efficient real-time data processing grounded by the underlying fundamental physics governing the system in focus.

\section{Conclusion}

In addressing the critical challenges associated with forecasting space weather phenomena like geomagnetic disturbances (GMDs) and geomagnetically induced currents (GICs) – the field has seen the emergence of Tiny Machine Learning (TinyML) as a tool to create ML-enabled magnetometer systems. Such systems predict terrestrial magnetic perturbations as a real-time proxy for GICs. Despite its promise in real-time data processing, TinyML is not without its constraints, primarily its intrinsic limitations preventing the use of more robust methods with higher computational requirements. This research introduced and developed a physics-guided TinyML framework to navigate these limitations. This framework adeptly incorporates regularization grounded in physical principles at pivotal junctures of the model's training and compression phases, consequently augmenting the precision and robustness of the resultant predictive outcomes. Fundamental to this approach is the adoption of a customized pruning strategy, utilizing the inherent physical characteristics of the domain to achieve an equilibrium between model compactness and the assurance of predictive robustness. The empirical findings of this study provide a thorough comparison between the developed physics-aware TinyML framework and its traditional counterparts. The comparative analysis presented herein highlights the enhanced accuracy and reliability of the proposed framework. It underscores its promising applicability in developing ML-enabled magnetometer systems for space weather prediction.

\section*{Acknowledgment}

We thank members of the the “Machine-learning Algorithms for Geomagnetically Induced Currents In Alaska and New Hampshire” (MAGICIAN) team. This work was supported by NSF EPSCoR Award OIA-1920965.

\bibliographystyle{IEEEtran}
\bibliography{ieee_phys_enhanced_TinyML}
\vspace{12pt}

\end{document}